\documentclass{article}
\usepackage{spconf,amsmath,graphicx}
\usepackage{amssymb}
\usepackage{multirow}

\title{Leveraging Local Temporal Information for Multimodal Scene Classification} 
\name{Saurabh Sahu, Palash Goyal}
\address{Samsung Research America}
\begin{document}
%
\maketitle
\begin{abstract}
Robust video scene classification models should capture the spatial (pixel-wise) and temporal (frame-wise) characteristics of a video effectively. Transformer models with self-attention which are designed to get contextualized representations for individual tokens given a sequence of tokens, are becoming increasingly popular in many computer vision tasks. However, the use of Transformer based models for video understanding is still relatively unexplored. Moreover, these models fail to exploit the strong temporal relationships between the neighboring video frames to get potent frame-level representations. In this paper, we propose a novel self-attention block that leverages both local and global temporal relationships between the video frames to obtain better contextualized representations for the individual frames. This enables the model to understand the video at various granularities. We illustrate the performance of our models on the large-scale YoutTube-8M data set on the task of video categorization and further analyze the results to showcase improvement.
\end{abstract}
\begin{keywords}
Video classification, self-attention, temporal information
\end{keywords}
\vspace{-3mm}
\section{Introduction}
\label{sec:intro}
The goal of video classification is to understand the visual and audio features to assign one or more relevant tags to the video. Such a task is relevant for applications such as content retrieval, mitigate missing video metadata, user profiling and content recommendation. 3D convolutional neural-nets (CNNs) have been used extensively for video classification tasks~\cite{feichtenhofer2020x3d, tran2019video, lin2019tsm}. However, their computational inefficiency has led the researchers to look for alternatives such as self-attention based Transformer architectures~\cite{vaswani2017attention}. Recently, ~\cite{dosovitskiy2020image} showed that Transformer based models train faster than their CNN counterparts, leading to enhanced scalability. Transformer based models also capture long-range dependencies which benefits many natural language processing (NLP) tasks ~\cite{xia2019tied, li2019beyond}. Capturing long-term dependencies across frames is crucial for video understanding as has been demonstrated by~\cite{wang2018non, wu2019long}. This has led to growing interest in using Transformer based models for video classification. 

Chen et al.~\cite{chen20182} proposed increasing the efficacy of convolutional blocks for action classification by using a two-stage attention mechanism to collect relevant features from different patches of video frames at various instants. Kmiec et al.~\cite{kmiec2018learnable} used Transformer encoder along with NetVLAD blocks to get effective multimodal feature representations for large scale video understanding.  Girdhar et al.~\cite{girdhar2019video} proposed ‘action Transformers’ where features extracted using 3D CNN are aggregated and fed to a self-attention block to leverage the spatio-temporal information around a person for action localization. Bertasius et al.~\cite{bertasius2021space} extended the idea of~\cite{dosovitskiy2020image} to propose Timesformer for video classification by using self-attention blocks to obtain spatiotemporal contextualized feature representations from a sequence of frame-level patches. These works show the ability of Transformer based models to generate video representations capturing the global context pertinent for the task at hand. While some of them exploit the spatial correlation between the pixels, none of them leverage the strong correlation between consecutive frames to obtain the contextualized output representations.

 We hypothesize that utilizing the strong temporal correlation existing within consecutive frames is crucial for video understanding. By focusing attention on just a portion of a video at any time, one can obtain more fine-grained information which can help with it's classification. Owing to the sequential nature of text, the importance of local attention maps in Transformer based models have been studied in NLP domain.~\cite{pande2020importance, child2019generating, beltagy2020longformer} have explored the importance of local attention maps for various language understanding tasks and proposed computationally efficient Transformer models. The local attention maps are created by enforcing a token to attend to other tokens only in a small neighbourhood around it by masking out tokens outside the neighborhood. In this paper, we explore various masks to compute different types local attention maps useful for video classification. We further propose to combine the global and local attention maps using gating mechanisms~\cite{shazeer2017outrageously}. Recent works have explored gating in Transformer based models~\cite{chai2020highway, fedus2021switch}, but they are not motivated by gating in attention space. Finally, we perform large scale experiments on the Youtube8M dataset~\cite{abu2016youtube} and show the advantage of our model over the baseline self-attention based model for video classification. We show that our approach can alleviate the effects of noisy frames in a video, by adding the information from local attention maps and correcting the baseline model's prediction.
\section{Methodology}
\label{sec:model}

Given audio $X_{a} \in \mathbb{R}^{T\times D_a}$ and visual $X_{v} \in \mathbb{R}^{T\times D_v}$ features  for $T$ frames of a video, we pass each of them through Transformer based architectures to obtain contextualized output representations $Y_{a} \in \mathbb{R}^{T\times D_a}$ and $Y_{v} \in \mathbb{R}^{T\times D_v}$ respectively. We then get a video level representation by concatenating the two contextualized representations and averaging across the time dimension. It is passed through a hidden layer before getting the final predictions. For our baseline Transformer based model, we use vanilla self-attention (followed by feed-forward and layer-norm operations as in ~\cite{vaswani2017attention}) to get the contextualized representations. After training the baseline model, we observe that the attention maps so generated do not capture the inter-frame correlations. Motivated by that, we propose various ways to compute local attention maps and use them with global attention maps for better video classification. In the following sections, we drop the subscript denoting modalities and note that similar Transformer architectures are used for both modalities.
\subsection{Baseline Model}
\label{sec:baseline}
In multi-headed self-attention with $M$ heads, input $X \in \mathbb{R}^{T\times D}$ is first transformed into query, key and value matrices. For $m$-th head, we compute the attention map $A_m \in \mathbb{R}^{T\times T}$ using the scaled dot product of the corresponding query $Q_m \in \mathbb{R}^{T\times D_M}$ and key $K_m \in \mathbb{R}^{T\times D_M}$ matrices ($D_m = D/M$). It is then multiplied with the value matrix $V_m \in \mathbb{R}^{T\times D_M}$ to get the corresponding head's output representation $O_{m} \in \mathbb{R}^{T\times D_M}$. These outputs from different heads are then concatenated and a linear transformation is done to get the final output representation $Y \in \mathbb{R}^{T\times D}$. 
\begin{eqnarray}
Q_m &=& XW^q_m, \hspace{1mm} K_m = XW^k_m, \hspace{1mm}V_m = XW^v_m \label{eqn:1}\\
A_m &=& \mbox{softmax} \big(\frac{Q_mK_m^T}{\sqrt{d_m}}\big)\label{eqn:2}\\
O_{m} &=& A_m.V_m \label{eqn:3}\\
Y &=& \mbox{concat}(O_{1},\ldots,O_{M}).W^o\label{eqn:4} 
\end{eqnarray}
\noindent where $W^q_m \in \mathbb{R}^{D\times D_M}$, $W^k_m \in \mathbb{R}^{D\times D_M}$, $W^v_m \in \mathbb{R}^{D\times D_M}$ and $W^o \in \mathbb{R}^{D\times D}$ are learnable matrices. Note that in this formulation, only a global attention map is computed by each head to get the final output representation. We call it global attention map as it depicts the importance of a frame $i$ with respect to frame $j$ taking into context the entire video.
\begin{figure}[t]
\hspace*{-3mm}
\includegraphics[scale=0.40]{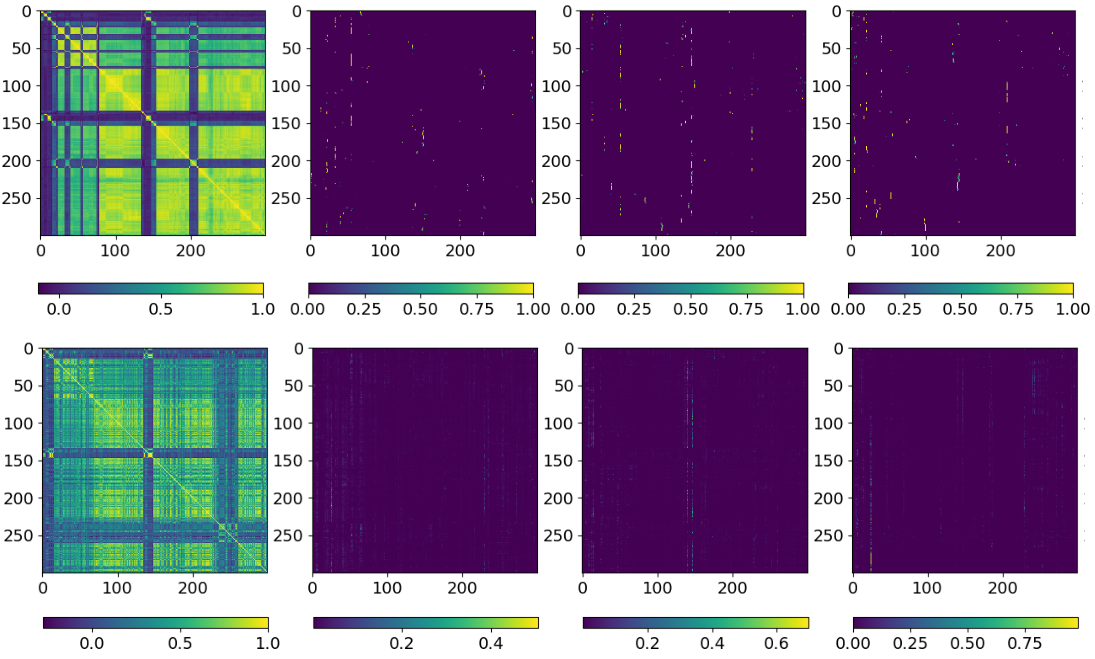}
\caption{Inter-frame similarity for visual (leftmost top) and audio features (leftmost bottom). Attention maps from three heads of the baseline self-attention models with visual (top right) and audio (bottom right) features as input.}
\label{fig:base_att}
\end{figure}
Once trained, we use the baseline model to compute the attention map for a video. We also compute the pair-wise cosine similarity matrix from the raw feature matrices and compare it with the attention maps. From Figure~\ref{fig:base_att}, we observe that the similarity matrices (especially the one computed from visual features) capture the high inter-frame correlation present in the video. In contrast, the attention maps are quite sparse. The vertical lines in the attention maps show that only a few selective frames are being attended to get the contextualized output representations. This can lead to incomplete predictions if high attention is given to only a handful of relevant frames, and incorrect predictions if given to irrelevant frame(s). Hence, we propose ways to leverage local attention maps for robust video classification.
\subsection{Leveraging local attention maps}
\label{sec:localatt}
To capture local information present in a specific segment of a video, we compute local attention maps. Using masks $mask_m \in \mathbb{R}^{T\times T}$, we can constrain a head to generate local
attention maps by modifying equation~\ref{eqn:2} as shown below.
\begin{eqnarray}
A_m &=& \mbox{softmax} \big(\frac{Q_mK_m^T}{\sqrt{d_m}} \odot mask_m\big) \label{eqn:5}\\
mask_m[i,j] &=&
\begin{cases}
    1,& \text{if } j \in N_i\\
    -\infty,              & \text{otherwise}
\end{cases} 
\end{eqnarray}
\begin{figure}[t]
\includegraphics[scale=0.20]{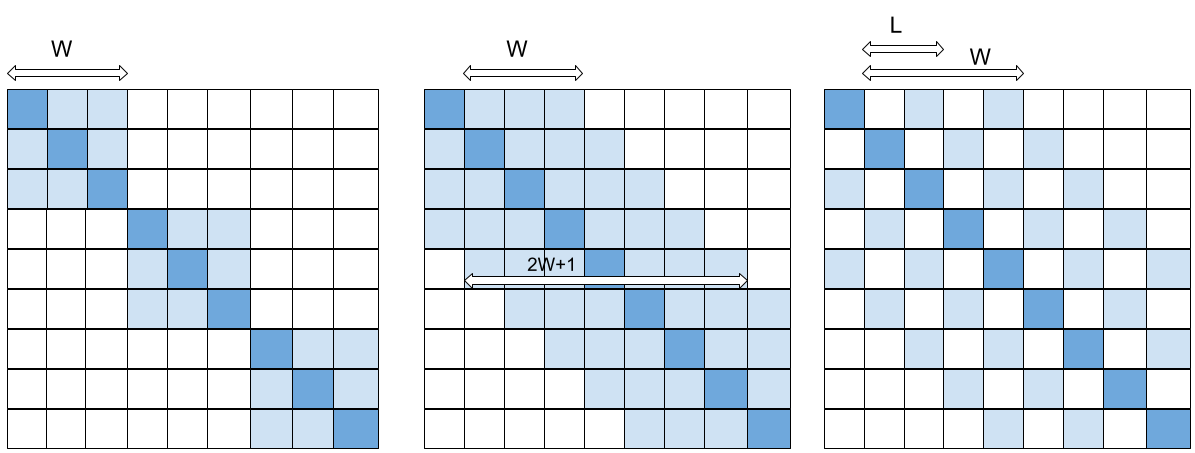}
\caption{$BD_W$ (left), $TP_W$ (center) and $TD_W^L$ (right) masks. Here, $W=3$ for $BD$ and $TP$ masks. For $TD$, $W=4$ and $L=2$. Dark blue entries depict the reference frames. Colored entries of the masks are 1 while rest are $-\infty$}
\label{fig:masks}
\end{figure}
$mask_m$ enforces a frame $i$ to attend to only the frames $j$ in its neighborhood $N_i$ by masking out other time-steps. We explore three different kind of masks shown in Figure~\ref{fig:masks}. Block-diagonal mask $BD_W$ effectively divides the video into $T/W$ non-overlapping segments and computes the local attention map for those segments. Toeplitz mask $TP_W$ defines a neighborhood of size $2W+1$ for a frame and computes the local attention map. Toeplitz-Dilated mask $TD_W^L$ forces a particular frame to focus on every $L$-th frame within a neighborhood of size $2W+1$.

Hence, a specific head's output $O_m$ obtained using equations~\ref{eqn:5} and~\ref{eqn:3} takes into context the inter-frame correlations occurring in a local neighborhood. A simple way to verify if local temporal information benefits our task is by having half the heads ($M/2$) generate global attention maps while the rest generate local attention maps. Final output representation is obtained using equations~\ref{eqn:3} and ~\ref{eqn:4} which then undergoes further operations same as the baseline model to get the final predictions. We describe more specifics about the architecture used in section~\ref{sec:expts}. Note that, we use {\fontfamily{qcr}\selectfont masked\_fill} function in Pytorch~\cite{NEURIPS2019_9015} instead of element-wise multiplication as depicted in equation~\ref{eqn:5} to prevent unstable gradients while backpropagation due to the $-\infty$ entries in the masks.
\subsection{Gating local and global contextual information}
\label{sec:gateatt}
The local and global contextual information can also be fused by using gating mechanisms. Unlike section~\ref{sec:localatt}, here we employ each of the $M$ heads to compute global and local attention maps. We propose two formulations depending on gating module implementation. In the following sections, we denote the global attention map computed using equation~\ref{eqn:2} as $A_m^g$ and the local attention map computed using equation~\ref{eqn:5} as $A_m^l$.
\subsubsection{Gating attention maps}
\label{sec:gating1}
Once $A_m^g$ and $A_m^l$ are computed, we use a gating network to get the resultant attention map $A_m$. We get the final output representation using equations~\ref{eqn:3} and~\ref{eqn:4} followed by the same operations as the baseline model to get the final predictions
\begin{eqnarray}
R^g, R^l &=& softmax([A_m^gW^g_g,A_m^lW^l_g]) \label{eqn:7} \\
A_m &=& R^g \odot A_m^g + R^l \odot A_m^l \label{eqn:8}
\end{eqnarray}
where $W^g_g, W^l_g \in \mathbb{R}^{T\times T}$ are learnable layers shared across the heads. We compute $A_m^gW^g_g$ and $A_m^lW^l_g$ followed by stacking the two resultant $T\times T$ matrices along a new dimension. Note that, $softmax$ is computed across this new dimension. It can be easily verified that $A_m$ from equation~\ref{eqn:8} is a valid attention map in the sense each row has non-negative elements that sum to one. 
\subsubsection{Gating contextual representations}
\label{sec:gating2}
Instead of gating at attention level, we can employ the gating mechanism to fuse information at a later stage. The representations obtained by multiplying the attention maps with the value matrix provide more contextual information than just the attention maps. Hence, fusing them using a gating mechanism could be more beneficial. Specifically, once $A_m^g$ and $A_m^l$ are computed, we compute the global and local contextual representations $O^g$ and $O^l$. Finally, we get the output representation $Y \in \mathbb{R}^{T\times D}$ which captures both local and global contextual information. This is passed through further layers same as the baseline model to get the final predictions.
\begin{eqnarray}
O^g &=& \mbox{concat}(A_1^gV_1,\ldots,A_M^gV_M) \label{eqn:9}\\
O^l &=& \mbox{concat}(A_1^lV_1,\ldots,A_M^lV_M) \label{eqn:10}\\
Y^g &=& O^gW^{og}, \hspace{2mm} Y^l = O^lW^{ol} \label{eqn:11}\\
R^g, R^l &=& softmax([O^gW^g_g,O^lW^l_g]) \label{eqn:12} \\
Y &=& R^g \odot Y^g + R^l \odot Y^l \label{eqn:13}
\end{eqnarray}
where $W^{og}, W^{ol}, W^g_g, W^l_g,  \in \mathbb{R}^{D\times D}$ are learnable layers.

\section{EXPERIMENTS AND RESULTS}
\label{sec:expts}
We use YouTube-8M dataset for our experiments which provides us with frame-wise video and audio features extracted at a rate of 1Hz using Inception v3 and VGGish respectively for 3862 classes~\cite{abu2016youtube}. We use binary cross-entropy loss to train our models. We evaluate our models using the metrics mentioned in ~\cite{lee20182nd}: (i) Global Average Precision (GAP), (ii) Mean Average Precision (MAP), (iii) Precision at Equal Recall Rate and (iv) Hit@1. 
Our training set consists of approximately 4 million videos. We use 32000 videos from the official development set for validation and use the rest as test set. We used Adam optimizer, with an initial learning rate of 0.0002 and batch size of 64. We compute validation set GAP every 10000 iterations and perform early-stopping with patience of 5.
Based on that, the learning-rate scheduler decreases the learning rate by a factor of 0.1 with patience of 3. We compare the following models with baseline:
\newline \textbf{ShareAtt:} We use 4 heads to generate local attention maps and 4 for global attention maps as described in section~\ref{sec:localatt}. For $BD$ we use the masks \{$BD_{10}, BD_{20}, BD_{30}, BD_{60}$\}, for $TP$ we use \{$TP_{10}, TP_{30}, TP_{60}, TP_{80}$\} and for $TD$ we use \{$TD_{10}^2, TD_{30}^3, TD_{60}^4, TD_{80}^5$\}.
\newline \textbf{GateAtt:} Gating mechanism described in section~\ref{sec:gating1}. All the heads use the same mask to compute local attention.
\newline \textbf{GateOp:} Gating mechanism described in section~\ref{sec:gating2}. All the heads use the same mask to compute local attention.
\newline For any specific mask-type used in GateAtt and GateOp models, we did not see a lot of performance variation when the parameters $W$ and/or $L$ were changed. We present the best results in Table~\ref{tab:results}. 
\begin{table*}[ht]
\centering
\caption{Comparing baseline model's performance with novel models on test set}
\begin{tabular}{c|c|c c c|c c c|c c c|}
\cline{2-11}
&Baseline & \multicolumn{3}{|c|}{ShareAtt} & \multicolumn{3}{|c|}{GateAtt} & \multicolumn{3}{|c|}{GateOp}\\
\cline{3-11}
&& $BD$  & $TP$  & $TD$  & $BD_{10}$  & $TP_{10}$  & $TD_{80}^{5}$  & $BD_{20}$  & $TP_{30}$  & $TD_{60}^{4}$  \\\hline
GAP & 85.07 & 85.18 & 85.25 & 85.16 & 85.21 & 85.30 & 85.30 & \textbf{86.03} & \textbf{86.13} & \textbf{85.99} \\
MAP & 44.61 & 44.81 & 45.04 & 44.69 & 44.89 & 45.28 & 45.31 & \textbf{47.03} & \textbf{47.49} & \textbf{46.98}\\
PERR & 78.97 & 79.05 & 79.22 & 79.03  & 79.13 & 79.27 & 79.21 & \textbf{79.96} & \textbf{80.10} & \textbf{79.88}\\
Hit@1 & 87.75 & 87.79 & 87.92 & 87.78  & 87.85 & 87.96 & 87.91 & \textbf{88.40} & \textbf{88.49} & \textbf{88.33}\\\hline
\end{tabular} 
\label{tab:results}
\end{table*}
\subsection{Results}
From Table~\ref{tab:results}, we see that our proposed novelties perform better than the baseline model. Improvement in MAP is more compared to GAP and Hit$@$1 suggesting that the performance boost is more for underrepresented classes. We observe that GateAtt models perform slightly better than the ShareAtt. This could be because we have more heads in GateAtt models to compute the attention maps than in ShareAtt models ($M$ instead of $M/2$). GateOp models perform significantly better than the other two models. This highlights the benefit of gating the contextualised representations obtained using global and local attentions rather than gating the attention maps themselves. Furthermore, for each of the novel models, we observe that $TP$ masks perform slightly better than the $BD$ mask. From Figure~\ref{fig:masks}, we observe that for any specific frame, $TP$ masks focus on equal number of past and future frames unlike the $BD$ masks thus providing the model with more elaborate local context. 
\subsection{Qualitative Analysis}
\begin{figure}[ht]
\includegraphics[scale=0.19]{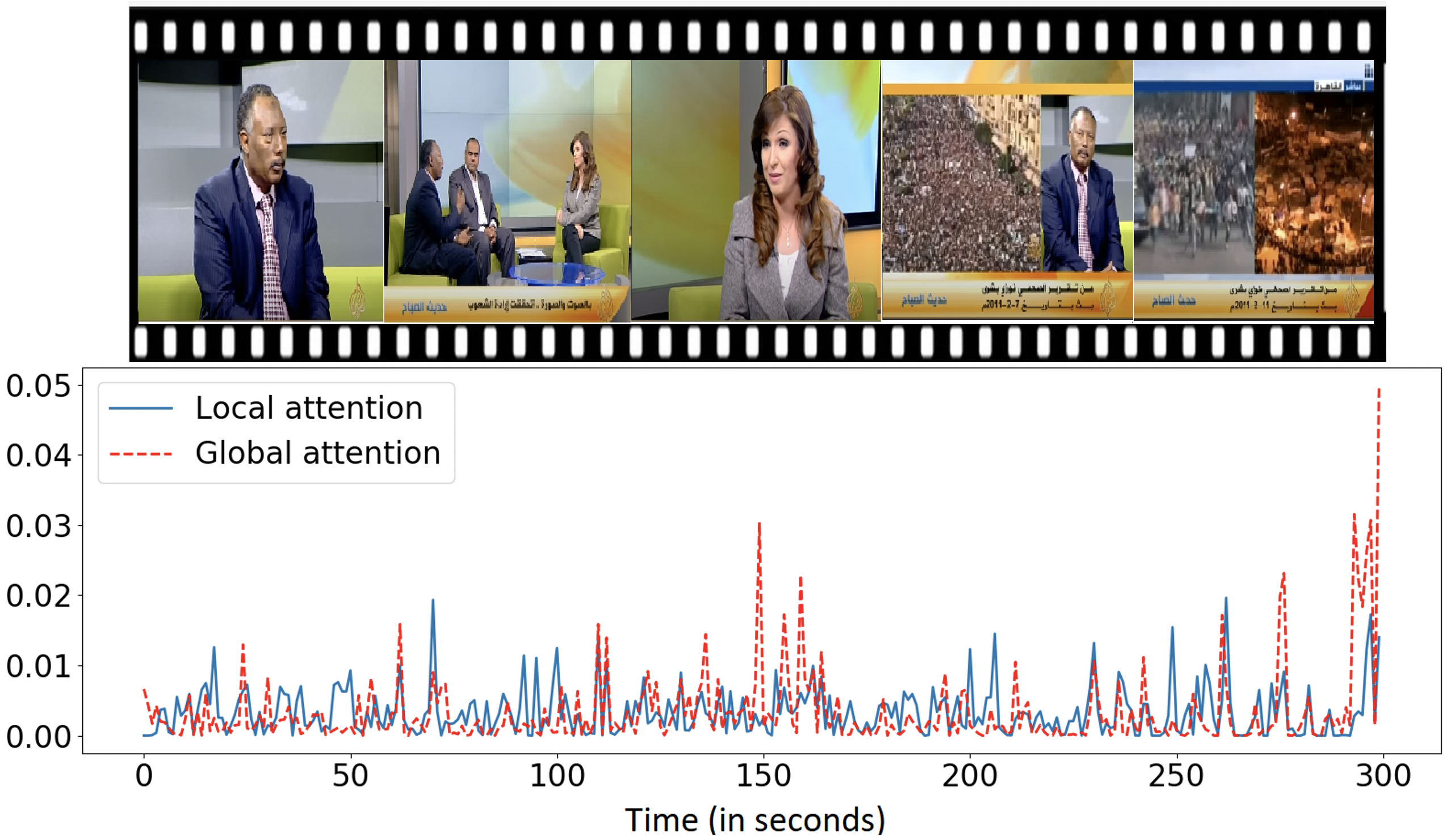}
\caption{Local and global attention profiles obtained for a video in test set. Link: \textit{youtube.com/watch?v=ygORXiV2Zpw}}
\label{fig:quality}
\vspace{-4mm}
\end{figure}

In Figure~\ref{fig:quality}, we show the local and global attention profiles obtained for a test-set video tagged as `News Program'. These profiles are obtained by averaging the attention maps over heads followed by averaging along the rows. We note that global attention profile is more uneven and puts most attention towards the end of the video showing a crowd of people. Hence, the baseline model predicts the video incorrectly (`Association Football'). In contrast, local attention profile is more uniform temporally. Using information from both local and global contexts, we get the correct prediction using GateOp model with $BD_{20}$ mask (`News Program').
\subsection{Importance of local information}
To analyze the importance of local contextual information, we computed the gradient matrix $G \in \mathbb{R}^{T\times T}$ using the trained models for 500 test set videos. $G[i,j]$ denotes the norm of the gradient of output frame $Y[i]$ with respect to input frame $X[j]$. We define a neighborhood $N_i$ for each frame $i$ and compute the following statistic that captures the sensitivity of that output frame to the local input frames.
\begin{equation}
S_i = \frac{avg_{j \in N_i}G[i,j]}{avg_{j \in N_i}G[i,j] + avg_{j \not \in N_i}G[i,j]}
\end{equation}
\begin{figure}[ht]
\hspace*{10mm}
\includegraphics[scale=0.30]{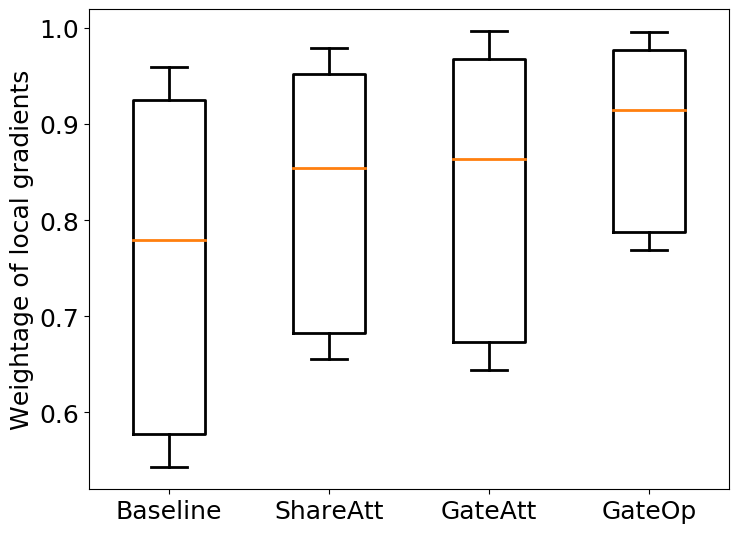}
\caption{Ratio of local to non-local gradients for visual features}
\label{fig:gradient}

\end{figure}
From Figure~\ref{fig:gradient}, it can be clearly seen that the output from the proposed novel models are more sensitive to the local inputs than baseline. Hence, taking into the account the results from Table~\ref{tab:results} we observe a positive correlation between model performance and sensitivity to local context.
\section{CONCLUSIONS}
\label{sec:conclusion}
In this paper, we proposed a novel way to incorporate local contextual information in self-attention models to improve video classification. We showed that the additional local context could mitigate the effects of incorrect global attention maps. Moreover, outputs of the better performing models were found to be more sensitive to local input frames than the baseline model. In future, we would like to explore video segmentation techniques to define better neighborhoods for computing local attention maps rather than a fixed-length mask based approach. We plan to extend this idea to include information from more than two levels of granularity by employing suitable masks and gating mechanisms.

\vfill\pagebreak



\bibliographystyle{IEEEbib}
\bibliography{paper}

\end{document}